\documentclass[a4paper, 10 pt, conference]{ieeeconf}  

\IEEEoverridecommandlockouts                              

\usepackage{hyperref}
\usepackage{url}
\usepackage{graphicx, subfigure}
\usepackage{cite}

\usepackage{amsmath}
\usepackage{amsfonts}

\usepackage{amssymb}
\newcommand{\ie}{\textit{i}.\textit{e}.}
\usepackage[ruled, linesnumbered]{algorithm2e}

\title{\LARGE \bf
Sample Efficient Reinforcement Learning via Model-Ensemble Exploration and Exploitation
}

\author{Yao Yao$^{1,2}$, Li Xiao$^{1}$, Zhicheng An$^{1,2}$, Wanpeng Zhang$^{1,2}$, and Dijun Luo$^{2}$
\thanks{$^{1}$TBSI, Tsinghua Shenzhen International Graduate School, Tsinghua University, Shenzhen, China {\tt\small \{y-yao19@mails, xiaoli@sz, azc19@mails, zhangwp19@mails\}.tsinghua.edu.cn}}%
\thanks{$^{2}$Tencent AI Lab, Shenzhen, China {\tt\small dijunluo@tencent.com}}%
}

\begin{document}

\maketitle
\thispagestyle{empty}
\pagestyle{empty}

\begin{abstract}
Model-based deep reinforcement learning has achieved success in various domains that require high sample efficiencies, such as Go and robotics. However, there are some remaining issues, such as planning efficient explorations to learn more accurate dynamic models, evaluating the uncertainty of the learned models, and more rational utilization of models. To mitigate these issues, we present MEEE, a model-ensemble method that consists of optimistic exploration and weighted exploitation. During exploration, unlike prior methods directly selecting the optimal action that maximizes the expected accumulative return, our agent first generates a set of action candidates and then seeks out the optimal action that takes both expected return and future observation novelty into account. During exploitation, different discounted weights are assigned to imagined transition tuples according to their model uncertainty respectively, which will prevent model predictive error propagation in agent training. Experiments on several challenging continuous control benchmark tasks demonstrated that our approach outperforms other model-free and model-based state-of-the-art methods, especially in sample complexity.
                               
\end{abstract}

\section{INTRODUCTION}

Reinforcement learning (RL) has achieved many remarkable results in recent years, including outperforming human performances on video games  \cite{mnih2015human,vinyals2019grandmaster,badia2020agent57}, mastering the game of GO \cite{silver2018general}, playing card games \cite{zha2019rlcard}, and even learning manipulation skills for robotics \cite{chatzilygeroudis2019survey,andrychowicz2020learning}. RL methods used in these domains are commonly divided into two categories: model-based RL and model-free RL. Model-based RL methods build a predictive model of the environment to assist the optimization of a policy or a controller, while in contrast model-free methods learn directly from agent’s interactions with the environment. Model-free RL has shown to obtain better asymptotic performance in the face of sequential decision-making problems, while the successes are highly sample inefficient \cite{kaiser2019model}. Model-based RL can significantly reduce the sample complexity by learning a model of the dynamics to do planning or generate simulations, which are substantially less expensive than interactions in a real environment. This property guarantees the potential of model-based RL in applications where sample efficiency plays a crucial role such as in robotics and automatic driving. Recently, significant improvements have been made in more sample-efficient model-based RL algorithms \cite{kurutach2018model,chua2018deep,janner2019trust,pathak2019self}. However, there is a substantial challenge for model-based RL, \ie, the model bias. With limited observation of the real environment, we can hardly build an accurate predictive model especially in high-dimensional tasks, typically resulting in performance degradation of the learned policy \cite{asadi2019combating}. This is because the model bias can be propagated and lead to an increase in overall error during agent training. 

There are two sources of model bias: aleatoric uncertainty and epistemic uncertainty \cite{chua2018deep,kong2020sde}. Concretely, aleatoric uncertainty is due to stochasticity of the dynamics or random noise during interactions, while epistemic uncertainty results from the lack of sufficient training data. One way to address the model bias is to use an ensemble of predictive models. One example is PETS \cite{chua2018deep} which was proposed to do trajectory optimization by introducing a ensemble of probabilistic models. ME-TRPO \cite{kurutach2018model} leveraged an ensemble of models to maintain the model uncertainty and regularize the learning process by policy validation.  Pathak, et al. \cite{pathak2019self} further proposed a method to support more effective exploration via model disagreement based on uncertainty estimates from ensembles. However, most prior researches have used ensemble methods to reduce model bias or guide exploration in isolation.


In this paper, we present MEEE, an ensemble method that is compatible with most of Dyna-style \cite{sutton1991dyna} RL algorithms. MEEE consists of the following key ingredients:
\begin{itemize}
    \item \textbf{Random initialization:} We initialize the model parameters randomly in order to enforce diversity between ensemble dynamic models, which stabilizes the learning process and improves performance.
    \item \textbf{Optimism-based exploration:} To further reduce the sample complexity and guide exploration in a more efficient manner, we propose a formulation for exploration inspired by the work in active learning literature \cite{shyam2019model} and upper-confidence bound method \cite{auer2002finite}. Our method selects optimal actions from an action candidates set and encourage exploration by adding an extra bonus for unconversant state-action space, where the model ensemble generates uncertain predictions.
    \item \textbf{Uncertainty-weighted exploitation:} Model bias can be exploited and propagated during agent training, which would lead to a deviation between the learned policy under the model ensemble and the true optimal policy under the real dynamic. To handle this issue, we assign different penalty coefficients to the augmented samples according to their uncertainty estimates under the model ensemble. Specifically, for those imagined samples with high uncertainty, the update of parameters is supposed to be more cautious, which can significantly mitigate error propagation when combining model-free methods with model-augmented transition samples.
\end{itemize}

We demonstrate the effectiveness of MEEE for continuous control benchmarks, \ie, OpenAI Gym \cite{brockman2016openai} and MuJoCo \cite{todorov2012mujoco}. In our experiments, MEEE consistently improves the performance of state-of-the-art RL methods and outperforms baselines, including SAC \cite{haarnoja2018soft} and MBPO \cite{janner2019trust}.

\section{Related Work} RL algorithms are well-known for their ability to acquire behaviors through trial-and-error with the environment \cite{sutton2018reinforcement}. However, such online data collection usually requires high sample complexity, which limits the flexibility of RL algorithms when applied in real-world applications. Although off-policy algorithms \cite{jiang2016doubly,haarnoja2018soft,fujimoto2019off} can in principle be applied to an offline setting \cite{lange2012batch}, they perform poorly in practice due to the limit of generalization to unseen states \cite{cobbe2019quantifying}, and can even pose risks in safety-critical settings \cite{garcia2015comprehensive}. To overcome these challenges, model-based RL presents an alternate set of approaches involving the learning of approximate dynamics models which can subsequently be used for policy search. Existing work on model-based RL utilizes generic priors like smoothness and physics \cite{denil2016learning,zeng2020tossingbot} for model learning, or directly models the dynamics by Gaussian processes \cite{deisenroth2011pilco}, local linear models \cite{kumar2016optimal}, and neural network function approximators \cite{draeger1995model,gal2016improving}. With a learned model, we are able to overcome the exploration-exploitation dilemma in RL.

\subsection{Model-based exploration}
Since the dynamics prediction model can reflect the agent’s capability of predicting the consequences of its behavior, the prediction error can be utilized to provide intrinsic exploration rewards (\emph{curiosity})~\cite{schmidhuber1991possibility}. The intuition behind this is that the higher the prediction error, the less familiar the predictive model is with that state-action pair. And this can be further inferred that the faster the error rate drops, the faster the model learning progress will be~\cite{weng2020exploration}.

The IAC~\cite{oudeyer2007intrinsic} approach outlines the idea of estimating learning progress and assigning intrinsic exploration bonuses through a forward dynamics prediction model. Considering the poor behavior when training the dynamics prediction model directly on raw pixels, Stadie et al.~\cite{stadie2015incentivizing} estimated the exploration reward in the encoding space by means of an autoencoder. In contrast, ICM~\cite{pathak2017curiosity} utilizes a self-supervised inverse dynamics model to approximate the state-space encoding function, robust to the uncontrollability of the environment. In addition to the aforementioned modeling approaches, the environment dynamics can also be modeled by variational inference. By doing so, VIME~\cite{houthooft2016vime} was proposed to maximize the information gain in the learning process of the prediction model, measured by the reduction of entropy.
In order to overcome the bias of one single predictive model, Pathak et al.~\cite{pathak2019self} proposed to use an ensemble of models to model the environment dynamics and calculated the disagreement among different model outputs as the additional exploration bonus.

\subsection{Model-based exploitation}
Model-based RL algorithm is a powerful tool to reduce the sample complexity by doing simulation or planning with the learned prediction model, which means that it is a good solution to the problem of insufficient and costly samples. PILCO \cite{deisenroth2011pilco} learns a probabilistic model through Gaussian process regression, which can express the uncertainty of environment well and make the model-based method more adaptable to complex environments. Deep-PILCO \cite{gal2016improving} generalizes to more complex environments by introducing Bayesian neural network (BNN) with high-capacity function approximators for neural networks. Model ensemble methods are further introduced to capture the uncertainty of predictions more comprehensively, which can enhance the explanatory power of the model and improve the robustness of learning policies.

By modularly integrating the environment models into model-free method, the sample efficiency of the original model-free algorithm can be significant improved. For example, Dyna-style methods \cite{sutton1991dyna,kurutach2018model, janner2019trust} use the models to generate augmented samples. PETS \cite{chua2018deep} directly uses the models for planning instead of performing explicit policy learning. However, all of these model-based methods suffer from model bias, and the performance degrades with the compounding error, resulting in bad decision policies. MVE \cite{feinberg2018model} and STEVE \cite{buckman2018sample} try to solve this problem by using short-horizon model-based rollouts and incorporating data from these rollouts into value estimation. 

\section{Background}
\label{bg}

We consider a Markov decision process (MDP \cite{bellman1957markovian}), defined by $\left(\mathcal{S}, \mathcal{A}, \mathcal{P}, \mathcal{R}, \gamma, \rho\right)$. Here $\mathcal{S} \subseteq \mathbb{R}^{n}$ and $\mathcal{A} \subseteq \mathbb{R}^{m}$ stand for the state and action spaces respectively, $\mathcal{P}: \mathcal{S} \times \mathcal{A} \rightarrow \mathcal{S}$ is a deterministic transition function to depict the dynamic, $\mathcal{R}: \mathcal{S} \times \mathcal{A} \rightarrow \mathbb{R}$ is a bounded reward function, $\gamma \in(0,1)$ is the discount factor, and $\rho$ is the initial state distribution. The agent would select an action $a_{t}$ according to a policy $\pi_{\theta}: \mathcal{S} \rightarrow \mathcal{A}$, while the initial state $s_{0} \sim \rho$. This generates a trajectory of states, actions, and rewards $\tau=\left(s_{0}, a_{0}, r_{0}, s_{1}, a_{1}, \ldots\right)$, where $a_{t} \sim \pi_{\theta}\left(. \mid s_{t}\right)$, $r_{t}=\mathcal{R}\left(s_{t}, a_{t}\right)$, and $s_{t+1}=\mathcal{P}\left(s_{t}, a_{t}\right)$. The goal of reinforcement learning is to find the optimal policy $\pi^{*}$ that maximizes the cumulative expected return, denoted by $\eta(\theta)$:
\begin{equation}
\pi^{*}=\underset{\pi}{\operatorname{argmax}}\ \eta(\theta)=\underset{\pi}{\operatorname{argmax}}\ E_{\tau}\left[\sum_{t=0}^{+\infty} \gamma^{t} r_{t}\right]
\end{equation}

Usually, model-based reinforcement learning methods assume the dynamic $\mathcal{P}$ and reward function $\mathcal{R}$ are unknown, and aim to construct a model $\mathcal{P}_{\phi}\left(s_{t+1}, r_{t}|s_{t}, a_{t}\right)$ to approximate the transition distribution and reward function by collecting interaction data with the true dynamics.

\begin{figure*}[htb]
\begin{center}
\begin{minipage}{1\textwidth}
\centering
\includegraphics[width=0.8\textwidth]{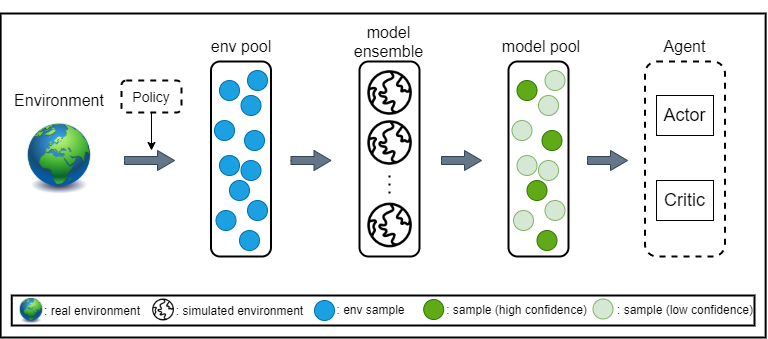}
\end{minipage}
\end{center}
\caption{Overview of the MEEE algorithm. By adopting optimism-based exploration, we continuously collect samples into the environment replay buffer $B_{env}$, which are used for training a set of BNN models. Then we sample from the model ensemble $\mathcal{P}_{\phi}$ to fill into a model replay buffer $B_{model}$, and assign different weights to these generated samples, based on which the agent is gradually optimized.} 
\label{fig_env}
\end{figure*}
\section{MEEE}
\label{gen_inst}

We present MEEE: \textbf{M}odel-\textbf{E}nsemble \textbf{E}xploration and \textbf{E}xploitation. Summarized generically in Algorithm \ref{algo:our-method}, MEEE 
executes exploratory actions to learn a more accurate ensemble model, optimizes the agent with imaginary data generated by the trained model, uses the updated policy with the ensemble model to evaluate the expected returns of the action candidates, then continues to determine the exploratory action iteratively. The full procedure is illustrated in Figure \ref{fig_env}. However, errors in the model predictions can be exploited during agent training, resulting in discrepancies between the optimal policy under the model and under the true environment. To mitigate this issue, we consider to assign discounted weights to the imagined data according to their uncertainty estimates. 


Instantiating MEEE amounts to specifying three design decisions: (1) the parameterization and training of the predictive model $\mathcal{P}_{\phi}$, (2) the optimization algorithm for the agent, and (3) how to take most advantages of the predictive model for high sample efficiency. 

\subsection{Model ensemble}
In our work, we set $\mathcal{P}_{\phi}$ (termed a \emph{model ensemble}) as an ensemble of dynamics models $\left\{p_{\phi}^{1}, \ldots, p_{\phi}^{I}\right\}$, where $p_{\phi}^{i}\left(.\mid s_{t}, a_{t}\right)= \mathcal{N}\left(\mu_{\phi}^{i}\left(s_{t}, a_{t}\right), \Sigma_{\phi}^{i}\left(s_{t}, a_{t}\right)\right)$ is a BNN which assumes the outputs obey a Gaussian distribution with diagonal covariance. The use of \emph{model ensemble} is a general setting in many researches \cite{houthooft2016vime,kurutach2018model,pathak2019self}, and is proven to be an efficient way to handle both of aleatoric uncertainty and epistemic uncertainty \cite{chua2018deep}. With different random initialization, we train each of the model separately based on the environment replay buffer $B_{env}$ via straightforward maximum likelihood estimation that minimizes the prediction error: 
\begin{equation}
 \mathcal{L}_{\phi}^{i}=\frac{\sum_{\tau_{t} \in B_{env}}\left[\left\|\left(s_{t+1},r_{t}\right)-\left(\hat{s}_{\left(t+1, i\right)}, \hat{r}_{\left(t, i\right)}\right)\right\|_{2}^{2}\right]}{|B_{env}|} 
\end{equation}
where $\tau_{t}=\left(s_{t}, a_{t}, r_{t}, s_{t+1}\right)$ is the ground truth, $\left(\hat{s}_{(t+1, i)}, \hat{r}_{(t, i)} \right) \sim p_{\phi}^{i}\left(s_{t}, a_{t}\right)$, and $i\in\left\{1,\cdots,I\right\}$. Standard techniques such as back-propagation through time (BPTT) and reparametrization trick \cite{kingma2015variational} are followed to facilitate the model learning.

\begin{algorithm*}[thp!]
\SetAlgoLined
Initialize policy $\pi_{\theta}$, Q-function $Q_{\zeta}$, predictive \emph{model ensemble} $\mathcal{P}_{\phi}$, augmentation function $f$, environment replay buffer $B_{env}$, and model replay buffer $B_{model}$ \\
\For{N epochs}{
    Train model $\mathcal{P}_{\phi}$ on $B_{env}$ via maximum likelihood \\
    \For{each step $t$}{
    Generate a base action according to the policy: $a_{t} \sim \pi_{\theta}\left(. \mid s_{t}\right)$ \\
    \tcp{OPTIMISM-BASED EXPLORATION}
    Collect $K$ action candidates using augmentation: $\left\{a_{t, i}=f(a_{t},i)\mid i\in[1 ; K]\right\}$ \\
    Choose the optimal action in the candidates set: $a^{*}_{t}=\arg\max _{a \in \{a_{t,i}\}}\left\{Q_{\zeta}\left(s_{t}, a\right)+\lambda \hat{V}\left(s_{t}, a\right)\right\}$ \\
    Store transition $\tau_{t}=\left(s_{t}, a^{*}_{t}, r_{t}, s_{t+1}\right)$: $B_{env}\leftarrow B_{env}\cup\{\tau_{t}\}$ \\
        \For{M model rollouts}{
            Sample $\hat{s}_{0}=s_{t} \sim B_{env}$ as an initial state \\
            \For{$i \in [0; k-1]$ steps}{
                Simulation in model $\mathcal{P}_{\phi}$ based on the policy:  $\hat{a}_{i} \sim \pi_{\theta}\left(. \mid \hat{s}_{i}\right)$ \\
                Store transitions and respective weights: $B_{model}\leftarrow B_{model}\cup\{\left(\hat{\tau}_{i}, w(\hat{\tau}_{i})\right)\}$ 
            }
         }
        \For{G gradient updates}{
            \tcp{UNCERTAINTY-WEIGHTED EXPLOITATION} 
            Update $Q_{\zeta}$ and $\pi_{\theta}$ by minimizing $\mathcal{L}_{\text {critic }}^{\text {MEEE }}(\zeta)$ and $\mathcal{L}_{\text {actor }}^{\text {MEEE }}(\theta)$ 
        }
    }
}
\caption{Model-Ensemble Exploration and Exploitation (\textbf{MEEE})}
\label{algo:our-method}
\end{algorithm*}

\subsection{Soft Actor-Critic}
Soft actor critic (SAC) algorithm \cite{haarnoja2018soft} is a state-of-the-art off-policy algorithm for continuous control problems. The training process of SAC parameters alternates between a soft policy evaluation and a soft policy improvement. At the soft policy evaluation step, SAC optimizes the critic by minimizing the Bellman residual, \ie, the critic loss:
\begin{equation}
\begin{aligned}
\mathcal{L}_{\text {critic }}^{\text {SAC }}(\zeta)&=\mathbb{E}_{\tau_{t} \sim B} \left[\mathcal{L}_{Q}\left(\tau_{t}, \zeta\right)\right],\\
\mathcal{L}_{Q}\left(\tau_{t}, \zeta\right)&=\left[Q_{\zeta}\left(s_{t}, a_{t}\right)-\bar{Q}_{t}\right]^{2}
\end{aligned}
\end{equation}
where $\tau_{t}$ is a transition sample, $B$ is a replay buffer, $\bar{Q}_{t}$ is the temporal difference (TD) target, and the Q-function parameterized by $\zeta$ is a neural network to approximate the cumulative expected return starting from $(s_t, a_t)$. With the delayed parameters $\bar{\zeta}$ and a temperature hyperparameter $\alpha$, the TD target can be calculated by the following equation: 
\begin{equation}
\bar{Q}_{t}=r_{t}+\gamma\mathbb{E}_{a_{t+1} \sim \pi_{\theta}}\left[\bar{Q}_{t+1}\right]
\end{equation}
where $\bar{Q}_{t+1}$ is modified by adding an entropy-based regularization term:
\begin{equation}
\bar{Q}_{t+1}=Q_{\bar{\zeta}}\left(s_{t+1}, a_{t+1}\right)-\alpha \log \pi_{\theta}\left(a_{t+1} \mid s_{t+1}\right)
\end{equation}

At the soft policy improvement step, SAC updates the policy $\pi_{\theta}$ by minimizing the actor loss:
\begin{equation}
\begin{aligned}
\mathcal{L}_{\text {actor }}^{\text {SAC }}(\theta)&=\mathbb{E}_{s_{t} \sim B}\left[\mathcal{L}_{\pi}\left(s_{t}, \theta\right)\right], \\
\mathcal{L}_{\pi}\left(s_{t}, \theta\right)&=\mathbb{E}_{a_{t} \sim \pi_{\theta}}\left[\alpha \log \pi_{\theta}\left(a_{t} \mid s_{t}\right)-Q_{\zeta}\left(s_{t}, a_{t}\right)\right]
\end{aligned}
\end{equation}
where the policy $\pi_{\theta}$ is modeled with a BNN. Our work updates the agent parameters by using the SAC algorithm in a Dyna \cite{sutton1991dyna} style, \ie, the replay buffer $B$ set to be the $B_{model}$, which collects transitions sampling from $\mathcal{P}_{\phi}$.

\subsection{Uncertainty quantification of model ensemble}
Model-based RL aims to find an optimal policy that achieves best expected return with high sample-efficiency. The key of sample efficiency lies in the learned model, but there exists one question: how confident is the model prediction? To answer this question, uncertainty quantification on state-action pairs is a prerequisite to evaluate the belief of the agent over the prediction model. Building upon the wealth of prior work \cite{strehl2008analysis,burda2018large,pathak2019self}, we develop and leverage proper uncertainty estimates that particularly suit the ensemble setting. Specifically, we leverage the variance over the \emph{model ensemble} to quantify the uncertainty in a self-supervised manner:
\begin{equation}
\begin{aligned}
\hat{V}\left(s_{t}, a_{t}\right) &\triangleq \operatorname{Var}\left(\left\{\mu_{\phi}^{i}\left(s_{t}, a_{t}\right) \mid i \in[1 ; I]\right\}\right) \\
&=\frac{1}{I-1} \sum_{i}\left(\mu_{\phi}^{i}\left(s_{t}, a_{t}\right)-\mu^{\prime}\right)^{2}
\end{aligned}
\end{equation}
where $\mu^{\prime}\triangleq\frac{1}{I} \sum_{i} \mu_{\phi}^{i}\left(s_{t}, a_{t}\right)$. High uncertainty indicates low confidence in prediction and thus requires more active exploration and conservative exploitation. As a result, there are two main usages of the uncertainty estimates: providing an additional bonus to encourage exploration and discounted weights to prevent model bias.

\begin{figure}[htbp]
\begin{center}
    \includegraphics[width=0.42\textwidth]{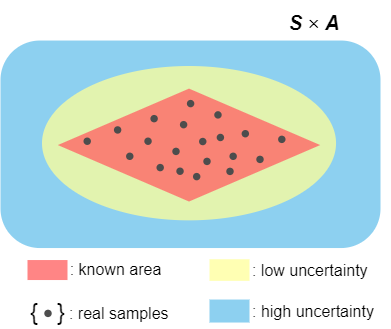}
\end{center}
\caption{Illustration of the uncertainty estimate, which partitions the state-action space into three parts with the support of real samples. As the number of samples in a certain area increases, we would learn more knowledge about that region. In order to obtain a greater information gain between the new belief over the prediction model to the old one, the agent is expected to take a new action in the uncertain region. 
}
\label{fig:ucb_2}
\end{figure}

\subsubsection{Optimism-based exploration}
Motivated by the upper confidence bound (UCB) algorithm \cite{auer2002finite}, we propose an exploration strategy to learn the predictive model more efficiently. Since $\hat{V}\left(s_{t}, a_{t}\right)$ can provide well-calibrated uncertainty estimates on unseen state-action pairs, our method adds an exploration bonus to expected return and chooses the action that maximizes: 
\begin{equation}
\label{exploration}
a_{t}=\arg\max _{a}\left\{Q_{\zeta}\left(s_{t}, a\right)+\lambda \hat{V}\left(s_{t}, a\right)\right\}
\end{equation}
where $\lambda>0$ is a hyperparameter, which defines the relative weighting of Q-function and the \emph{model ensemble} uncertainty. Our work differs from prior model-ensemble methods in that we consider both best estimates of Q-functions and uncertainty. The intuition is illustrated in Fig. \ref{fig:ucb_2}.  We remark that the theoretical connection between information gain and the \emph{model ensemble} uncertainty have already been discussed in \cite{houthooft2016vime,shyam2019model,sekar2020planning}. However, only maximizing the uncertainty is sample-inefficient as the agent would always seek to the novelist state space no matter how much the Q-fucntion is, which can incur risk actions. On the other hand, only Q-function values considered will limit the efficiency of exploration and result in slower model learning. To mitigate this problem, the optimism-based exploration is proposed to incentivize the policy to visit high-value state-action pairs, which are equipped with both large Q-functions and information gains. 

\begin{figure}[htbp]
\begin{center}
    \includegraphics[width=0.48\textwidth]{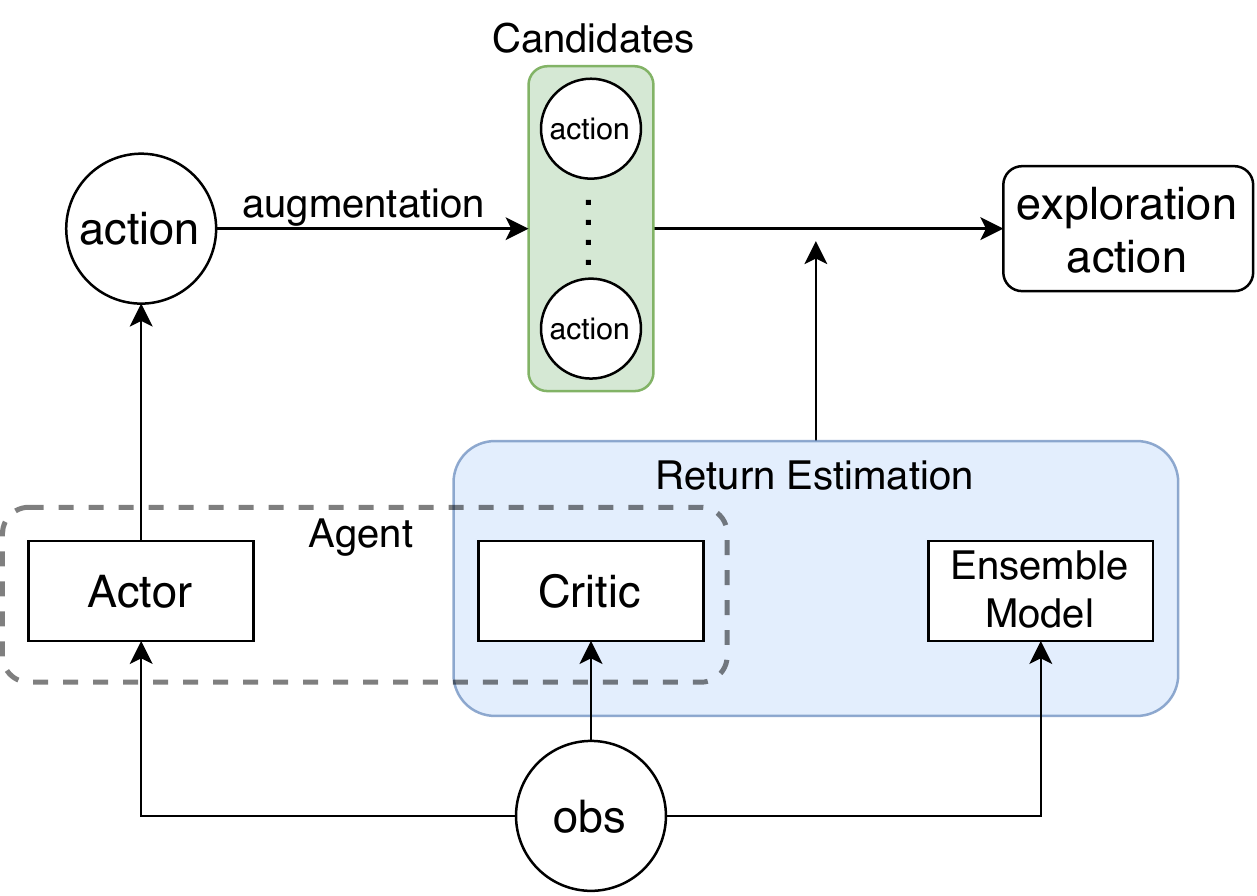}
\end{center}
\caption{Illustration of optimism-based exploration.}
\label{fig:ucb}
\end{figure}

Specifically, it is not straightforward to find the action that maximizes Eq.(\ref{exploration}) in continuous action spaces. To handle this issue, we propose a simple approximation scheme (see Fig. \ref{fig:ucb}), which generates $K$ action candidates by using $f$ to augment the action $a_{t} \sim \pi_{\theta}$, and then selects the action that maximizes Eq.(\ref{exploration}). Practically, our work uses random Gaussian noise generator for augmentation, \ie,  $f\left(a_t\right)=a_t+z, z \sim \mathcal{N}\left(\textbf{0},\Psi\right)$, where $\Psi>0$ is a hyperparameter to control the scope of the disturbance. We insist that other data augmentation methods can also be used in candidates generation in principle.

\subsubsection{Uncertainty-weighted exploitation}
\label{sec:weight}
Since conventional model-based methods in a Dyna \cite{sutton1991dyna} framework such as ME-TRPO \cite{kurutach2018model} and MBPO \cite{janner2019trust} utilize $B_{model}$, the compounding error inheriting from model-bias would propagate during agent training. This error propagation is shown to cause convergence and divergence between the optimal policies under the true dynamic $\mathcal{P}$ and the \emph{model-ensemble} $\mathcal{P}_{\phi}$ \cite{janner2019trust}. To mitigate this issue, we propose to assign different discount weights to the imagined samples based on the uncertainty estimates as follows:
\begin{equation}
w(\hat{\tau}_{t})=\sigma\left(-\hat{V}\left(\hat{s}_{t}, \hat{a}_{t}\right)  * T\right)+0.5
\end{equation}
where $T>0$ is another temperature hyperparameter, $\hat{\tau}_{t}=\left(\hat{s}_{t}, \hat{a}_{t}, \hat{r_{t}}, \hat{s}_{t+1}\right) \sim \mathcal{P}_{\phi}$ is an imagined transition, and $\sigma$ is the sigmoid function. Because $\hat{V}\left(\hat{s}_{t}, \hat{a}_{t}\right)$ is always nonnegative, the discounted weight is bounded in $\left[0.5, 1.0\right]$. This empirical setting of $w(\hat{\tau}_{t})$ is referred to SUNRISE \cite{lee2020sunrise}. 

In order to obtain a better capacity of resisting model bias, we down-weight the sample transitions with high uncertainty by modifying the proposed objectives of SAC as following:
\begin{equation}
\begin{aligned}
\mathcal{L}_{\text {critic }}^{\text {MEEE }}(\zeta) &=w(\hat{\tau}_{t}) * \mathcal{L}_{\text {critic }}^{\text {SAC }}(\zeta), \\
\mathcal{L}_{\text {actor }}^{\text {MEEE }}(\theta) &=w(\hat{\tau}_{t}) * \mathcal{L}_{\text {actor }}^{\text {SAC }}(\theta)
\end{aligned}
\end{equation}

Here, imagined transitions with lower confidence, \ie, greater uncertainty, would result in more discreet updates, which prevent the model bias propagation during agent training and are shown to provide performance guarantee for model-based RL algorithms. In principle, a potential benefit of a discounted weight is that the policy is still allowed to learn from the inaccurate imagined samples for a better generalization performance.

\begin{figure*}[htbp]
\centering
\subfigure[HalfCheetah-v2]{
\begin{minipage}[t]{0.23\textwidth}
\centering
\includegraphics[width=0.8\textwidth]{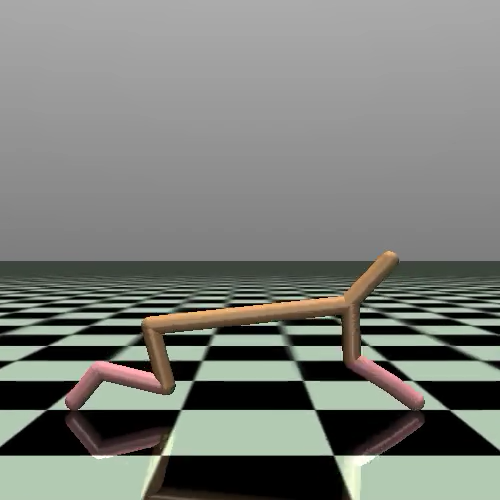}
\end{minipage}%
}%
\hfill
\subfigure[Ant-v2]{
\begin{minipage}[t]{0.23\textwidth}
\centering
\includegraphics[width=0.8\textwidth]{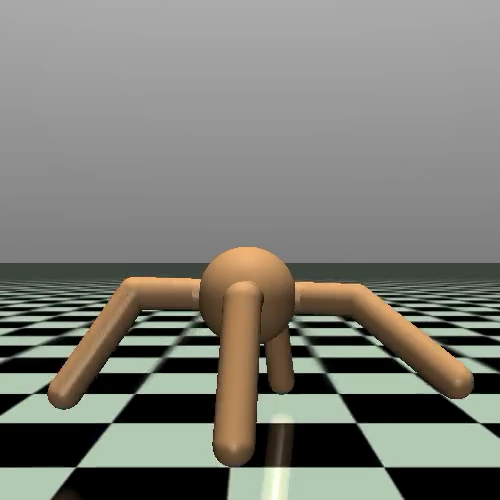}
\end{minipage}%
}%
\hfill
\subfigure[Humanoid-v2]{
\begin{minipage}[t]{0.23\textwidth}
\centering
\includegraphics[width=0.8\textwidth]{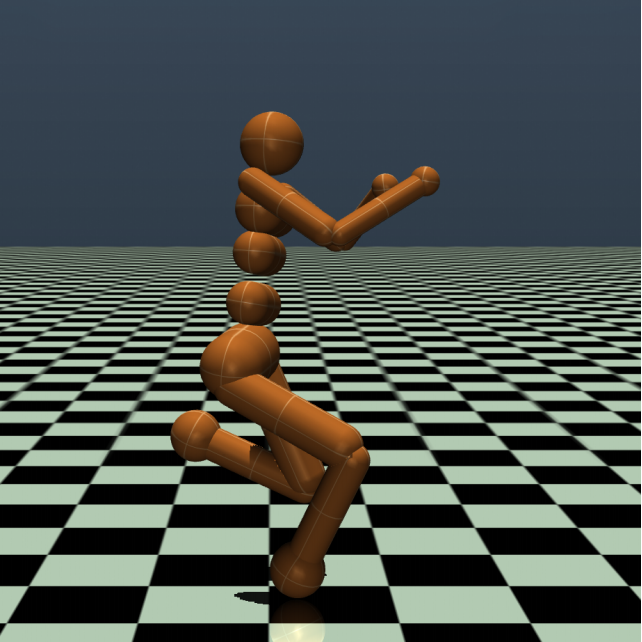}
\end{minipage}
}%
\hfill
\subfigure[Walker2d-v2]{
\begin{minipage}[t]{0.23\textwidth}
\centering
\includegraphics[width=0.8\textwidth]{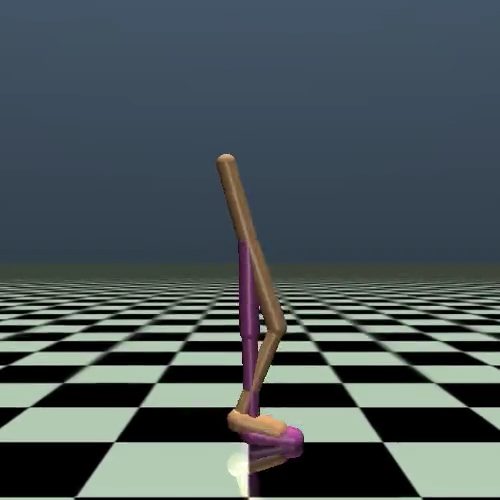}
\end{minipage}
}%
\centering
\caption{OpenAI Gym tasks used in our experiments. These tasks require the RL agent to learn locomotion gaits for the illustrated simulated characters.
}
\label{mujoco}
\end{figure*}

\section{Experimental results}
Our experiments aims to answer the following questions:
\begin{itemize}
\item \textbf{Comparison to prior work:} How does MEEE perform in common benchmark tasks in comparison to prior state-of-the-art RL algorithms?
\item \textbf{Contribution of each technique in MEEE:} What is the contribution of optimism-based exploration and uncertainty-weighted exploitation for MEEE?
\end{itemize}

To answer the posed questions, we consider four common benchmark tasks from OpenAI Gym \cite{brockman2016openai} simulated with MuJoCo \cite{todorov2012mujoco} physics engine: HalfCheetah-v2, Ant-v2, Humanoid-v2 and Walker2d-v2, which are illustrated in Fig. \ref{mujoco}. In our comparisons, we compare to SAC \cite{haarnoja2018soft} and MBPO \cite{janner2019trust}, which represent the state-of-the-art in both model-free and model-based RL. The experimental code is publicly available at
\url{https://github.com/YaoYao1995/MEEE}.

\begin{figure*}[!tbhp]
\centering
\subfigure[Comparison of SAC, MBPO and MEEE]{
\begin{minipage}[t]{0.47\textwidth}
\centering
\includegraphics[width=1\textwidth]{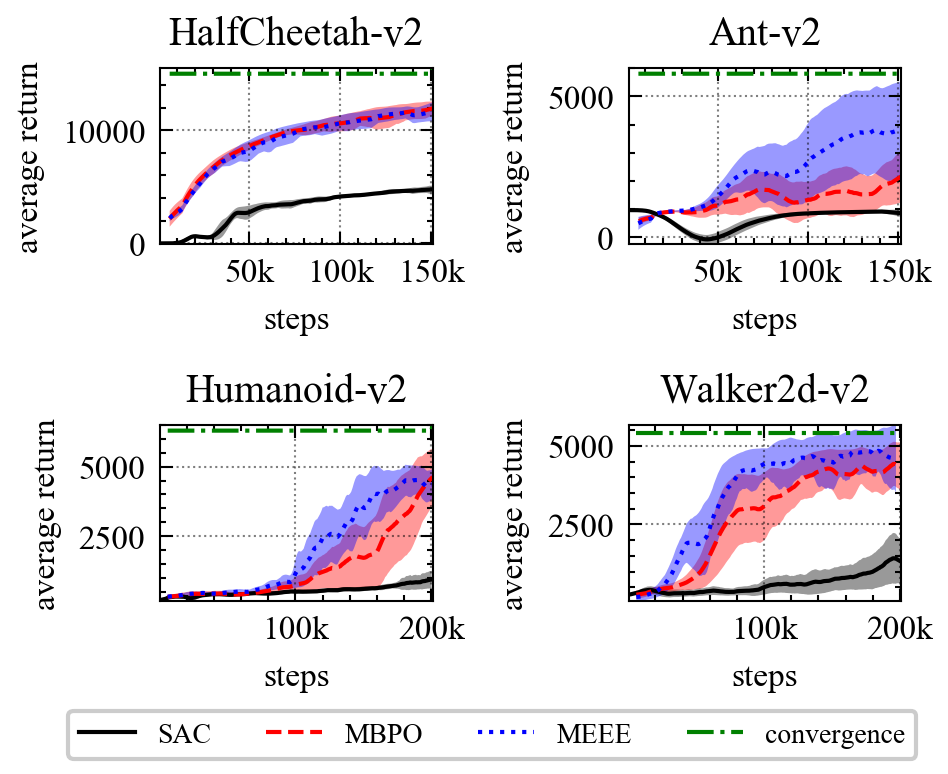}
\label{fig:result}
\end{minipage}%
}%
\hfill
\subfigure[Ablation experiments]{
\begin{minipage}[t]{0.47\textwidth}
\centering
\includegraphics[width=1\textwidth]{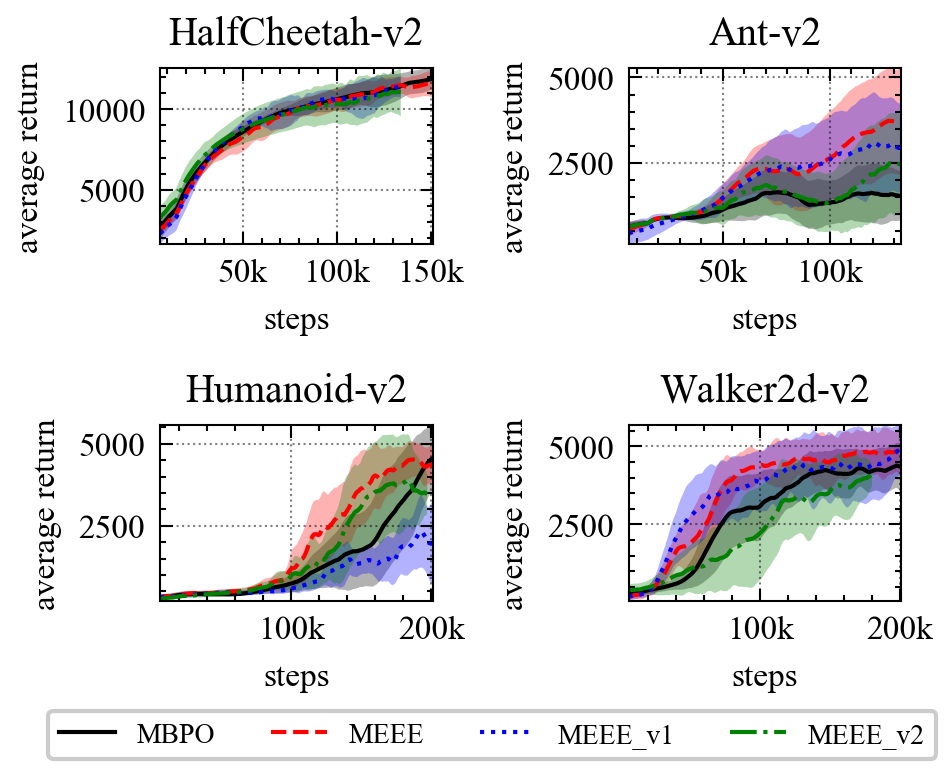}
\label{fig:ablation}
\end{minipage}%
}%
\centering
\caption{Learning curves of MEEE and corresponding baselines on four benchmark tasks. The x-axis plots the number of interactions with the MuJoCo environment, while the y-axis plots the average return. Colored curves indicate the mean of all trials with four different seeds and shaded regions correspond to standard deviation among trials, each of which is evaluated on the standard 1000-step versions of the benchmarks.} 
\end{figure*}

\subsection{Comparison with State-of-the-Arts}
Note that MEEE is a unified framework for utilizing uncertainty quantification to provide a better exploration–exploitation tradeoff, so MEEE can in principle be used in conjunction with most of the Dyna-style RL algorithms. In order to make a fair comparison, we implement MEEE by slightly modifying MBPO to support optimism-based exploration and uncertainty-weighted exploitation, and the additional hyperparameters are simply set as: 
$\lambda=1, \Psi=\mathbf{1}, T=20$. We declare that this is not the optimal setting after fussy parametric selection.

As shown in Fig. \ref{fig:result}, we plot learning curves for all methods, along with the asymptotic performance of SAC which does not converge in the shown region. These results highlight the strength of MEEE in terms of performance and sample complexity as MEEE outperforms other methods by a significant margin on three of the benchmark tasks. For example, MEEE’s performance on Walker2d-v2 at 200 thousand steps matches that of SAC at 2 million steps. However, notice that there is little difference in performance between MEEE and MBPO on HalfCheetah-v2. One possible reason is that the state space of HalfCheetah-v2 is less complex, so the exploration and exploitation mechanism of the vanilla MBPO is more than adequate. We believe future work on improving uncertainty estimation and careful hyperparameter tuning may be able to improve the performance further. 

\subsection{Ablation Study}
In order to verify the individual effects of each technique in MEEE, we conduct a thorough ablation study on MEEE. The main goal of the ablation study is to understand how uncertainty-weighted exploitation and optimism-based exploration affect performance. We denote MEEE\_v1 as the version only using uncertainty-weighted exploitation, and MEEE\_v2 to indicate only optimism-based exploration was considered. 

The results of our study are shown in Fig. \ref{fig:ablation}. Surprisingly, we find that the performance of MEEE\_v1 and MEEE\_v2 differs in different tasks. For example, MEEE\_v2 performs comparably to the vanilla MEEE on Humanoid-v2, while the performance on Walker2d-v2 is even worse than MBPO, possibly because Humanoid-v2 requires greater generalization and hence places more demands on effective exploration. On the contrary, we notice that MEEE\_v1 performs pretty well on Ant-v2 and Walker2d-v2 but performs poorly on Humanoid-v2, which suggests that uncertainty penalty is able to make major contributions when the model bias of the \emph{model ensemble} is relatively severe. 

To sum up, using uncertainty-weighted exploitation or optimism-based exploration alone sometimes degrades performance slightly. However, if they are used in a complementary way, MEEE will surely attain the improved performance, which is proven to be empirically successful.

\section{CONCLUSIONS}
In this work, we present a simple unified ensemble method namely MEEE that integrates random initialization, optimism-based exploration, and uncertainty-weighted exploitation to handle various issues in model-based RL algorithms. Our experiments show that MEEE consistently outperforms state-of-the-art RL algorithms on several benchmark continuous control tasks. We further evaluate the effect of each key component of our algorithm, showing that both optimism-based exploration and uncertainty-weighted exploitation are essential for successful applications of MEEE. For future work, we will investigate the integration of uncertainty quantification into other model-based RL frameworks and study how to make a better uncertainty estimation. Another enticing direction for future work would be the application of MEEE to real-world robotics systems.


\addtolength{\textheight}{-0.3cm}   







\bibliographystyle{IEEEtran} 
\bibliography{IEEEabrv,IEEEexample}

\end{document}